\documentclass[10pt,twocolumn,letterpaper]{article}

\usepackage{cvpr}              % To produce the CAMERA-READY version
\usepackage{graphicx}
\usepackage{amsmath}
\usepackage{amssymb}
\usepackage{booktabs,comment}

\usepackage{makecell}
\usepackage{multicol}

\usepackage{multirow}

\usepackage{pifont}

\newcommand{\cmark}{\ding{51}}%

\newcommand{\revised}{\textcolor{black}}%

\usepackage[pagebackref,breaklinks,colorlinks]{hyperref}

\usepackage[capitalize]{cleveref}
\crefname{section}{Sec.}{Secs.}
\Crefname{section}{Section}{Sections}
\Crefname{table}{Table}{Tables}
\crefname{table}{Tab.}{Tabs.}

\def\cvprPaperID{07678} % *** Enter the CVPR Paper ID here
\def\confName{CVPR}
\def\confYear{2022}

\begin{document}

%%%%%%%%% TITLE - PLEASE UPDATE
%\title{Unsupervised Mutually Cooperative Generative Discriminative Network for Unsupervised Video Anomaly Detection}

%\title{Mutually Cooperative Generative Discriminative Network for Unsupervised Video Anomaly Detection}

% \title{Lousy Mutually-Teaching Networks for Unsupervised Video Anomaly Detection}
% \title{Mutually Cooperative Learning for Video Anomaly Detection using Unlabelled Training Data}
% \title{Generative Friendly Networks for Unsupervised Video Anomaly Detection}
\title{Generative Cooperative Learning for Unsupervised Video Anomaly Detection}
%\title{Noisy-teacher Clean-student Network for Unsupervised Video Anomaly Detection }
%\title{Lousy-Teacher Efficient-Student Network for Unsupervised Video Anomaly Detection}  
\author{M. Zaigham Zaheer$^{1,2,3,5}$, Arif Mahmood$^{4}$, M. Haris Khan$^{5}$, Mattia Segù$^{3}$, Fisher Yu$^{3}$, Seung-Ik Lee$^{1,2}$\\
Electronics and Telecommunications Research Institute$^{1}$, Univ. of Science and Technology$^{2}$,\\ETH Zurich$^{3}$, Information Technology Univ.$^{4}$, Mohamed bin Zayed Univ. of Artificial Intelligence$^{5}$\\
% {\tt\small mzz@etri.re.kr, arif}
% For a paper whose authors are all at the same institution,
% omit the following lines up until the closing ``}''.
% Additional authors and addresses can be added with ``\and'',
% just like the second author.
% To save space, use either the email address or home page, not both
% \and
% Second Author\\
% Institution2\\
% First line of institution2 address\\
% {\tt\small secondauthor@i2.org}
}

\maketitle

\begin{abstract}
Video anomaly detection is well investigated in weakly-supervised and one-class classification (OCC) settings. However, unsupervised video anomaly detection methods are quite sparse, likely because anomalies are less frequent in occurrence and usually not well-defined, which when coupled with the absence of ground truth supervision, could adversely affect the performance of the learning algorithms. This problem is challenging yet rewarding as it can completely eradicate the costs of obtaining laborious annotations and enable such systems to be deployed without human intervention. To this end, we propose a novel unsupervised Generative Cooperative Learning (GCL) approach for video anomaly detection that exploits the low frequency of anomalies towards building a cross-supervision between a generator and a discriminator. In essence, both networks get trained in a cooperative fashion, thereby allowing unsupervised learning. We conduct extensive experiments on two large-scale video anomaly detection datasets, UCF crime and ShanghaiTech. Consistent improvement over the existing state-of-the-art unsupervised and OCC methods corroborate the effectiveness of our approach. 
\end{abstract}

%%%%%%%%% BODY TEXT

\section{Introduction}
\label{sec:intro}

\begin{figure}[t]
\begin{center}
    
       \includegraphics[width=1\linewidth]{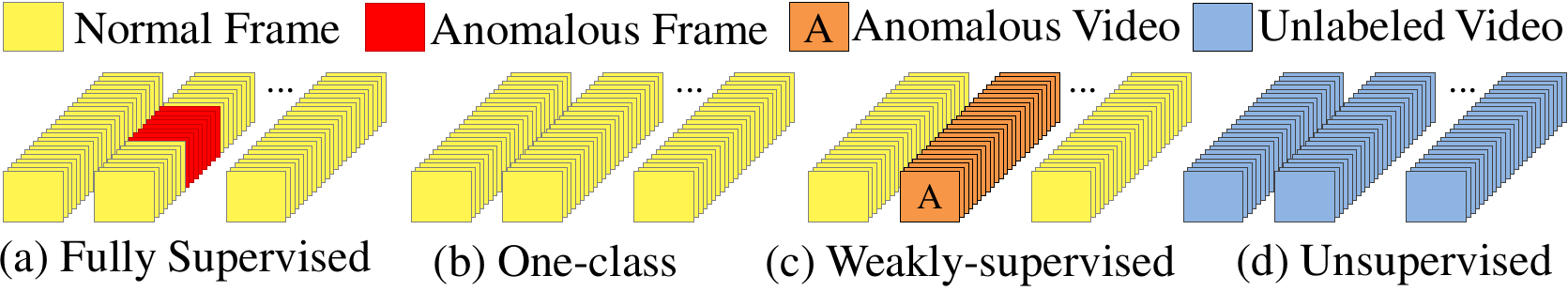}
       \vspace{-2.5em}

\end{center}
     \caption{Different training modes for video anomaly detection: (a) Fully supervised mode requires frame-level normal/abnormal annotations in the training data. (b) One-Class Classification (OCC) requires only normal training data. (c) Weakly supervised mode requires video-level normal/abnormal annotations.  (d) Unsupervised mode requires no training data annotations.}
     \label{fig:training_modes} \vspace{-1.5em}
   \end{figure}

In the real world, learning-based anomaly detection tasks are extremely challenging mainly because of the rare occurrence of such events. The challenge further exacerbates owing to the unconstrained nature of these events. Obtaining sufficient anomaly examples is thus quite cumbersome, while one may safely assume that an exhaustive set, particularly required for training fully-supervised models, will never be collected.
To make learning tractable, anomalies have often been attributed as significant deviations from the normal data. Therefore, a popular approach towards anomaly detection is to train a one-class classifier which learns the dominant data representations using only normal training examples \cite{zaheer2020old,liu2018future,zhang2016video,luo2017revisit,xia2015learning,hinami2017joint,sabokrou2017deep_novelty,smeureanu2017deep,ravanbakhsh2018plug,ravanbakhsh2017abnormal,hasan2016anomaly} (Fig. \ref{fig:training_modes}). 
A noticeable drawback of one-class classification (OCC) based methods is the limited availability of the normal training data, not capturing all the normalcy variations~\cite{chandola2009anomaly}. In addition, the OCC approaches are usually unsuitable for complex problems with diverse multiple classes and a wide range of dynamic situations often found in video surveillance. In such cases, an unseen normal activity may deviate significantly enough from the learned normal representations to be predicted as anomalous, resulting in more false alarms {\cite{hasan2016anomaly,zaheer2020claws,zaheer2020cleaning}}.

Recently, weakly supervised anomaly detection methods have gained significant popularity \cite{liu2019completeness,liu2019weakly,yu2019temporal,narayan20193c,shou2018autoloc,wang2017untrimmednets} that reduce the cost of obtaining manual fine-grained annotations by employing video-level labels
\cite{zaheer2020claws,zaheer2020cleaning,zaheer2020self,sultani2018real,zhong2019graph}. Specifically, 
a video is labeled  as anomalous if \textit{some} of its contents are anomalous and normal if \textit{all} of its contents are normal, requiring careful manual inspection of the full video contents.  Although such annotations are relatively cost-effective, yet remain impractical in many real-world  applications.
There is a plethora of video data, specifically raw footage, that can be leveraged for anomaly detection training if no annotation cost is incurred. Unfortunately, to the best of our knowledge, there are hardly any notable attempts in leveraging  the unlabelled training data for video anomaly detection.

In the current work, we explore \emph{unsupervised mode} for video anomaly detection that is certainly more challenging than fully, weakly or one-class supervision (Fig. \ref{fig:training_modes}). However, it is also more rewarding due to minimal assumptions and hence will encourage the development of novel and more practical algorithms. Note that, the term `unsupervised' in literature often refers to OCC approaches which assume all normal training data \cite{zaheer2020old,park2020learning,gong2019memorizing}. However, it renders the overall learning problem partially supervised \cite{jewell2021oled}.
In approaching unsupervised anomaly detection in surveillance videos, we exploit the simple facts that videos are information-rich compared to still images and anomalies are less frequent than the normal happenings  \cite{zaheer2020claws,sultani2010abnormal,chan2008ucsd,shanghaiTech2017}, and attempt to leverage such domain knowledge in a structured manner.

To this end, we propose a \emph{Generative Cooperative Learning (GCL)} method which takes \emph{unlabelled} videos as input and learns to predict frame-level anomaly score predictions as output. 
The proposed GCL comprises two key components, a \emph{generator} and a \emph{discriminator}, which essentially get trained in a mutually cooperative manner towards improving the anomaly detection performance. 
The generator not only reconstructs the abundantly available normal representations but also distorts the possible high-confidence anomalous representations by using a novel negative learning (NL) approach. 
The discriminator instead estimates the probability of an instance to be anomalous. Since we approach unsupervised anomaly detection, we create pseudo-labels from generator and use these to train the discriminator and in the following step, we create pseudo-labels from the trained version of discriminator and then use these to improve the generator. 
The overall system is trained in an alternate training fashion where, in each iteration, both the generator and the discriminator get improved with mutual cooperation. 

\noindent\textbf{Contributions.} 
We propose an anomaly detection approach capable of localizing anomalous events in complex surveillance scenarios without requiring labelled training data. To the best of our knowledge, our method is the first rigorous attempt tackling the surveillance videos anomaly detection in a fully unsupervised mode. A novel Generative Cooperative Learning (GCL) framework is proposed that comprises a generator, a discriminator, and cross-supervision. The generator network is forced not to reconstruct anomalies by using a novel negative learning approach. Extensive experiments on two large-scale complex anomalous event detection datasets, UCF-Crime and ShanghaiTech, show that our method provides visible gains over the baselines and several existing unsupervised as well as OCC methods.

\begin{figure*}[t]
  \centering
  \includegraphics[width=0.95\linewidth]{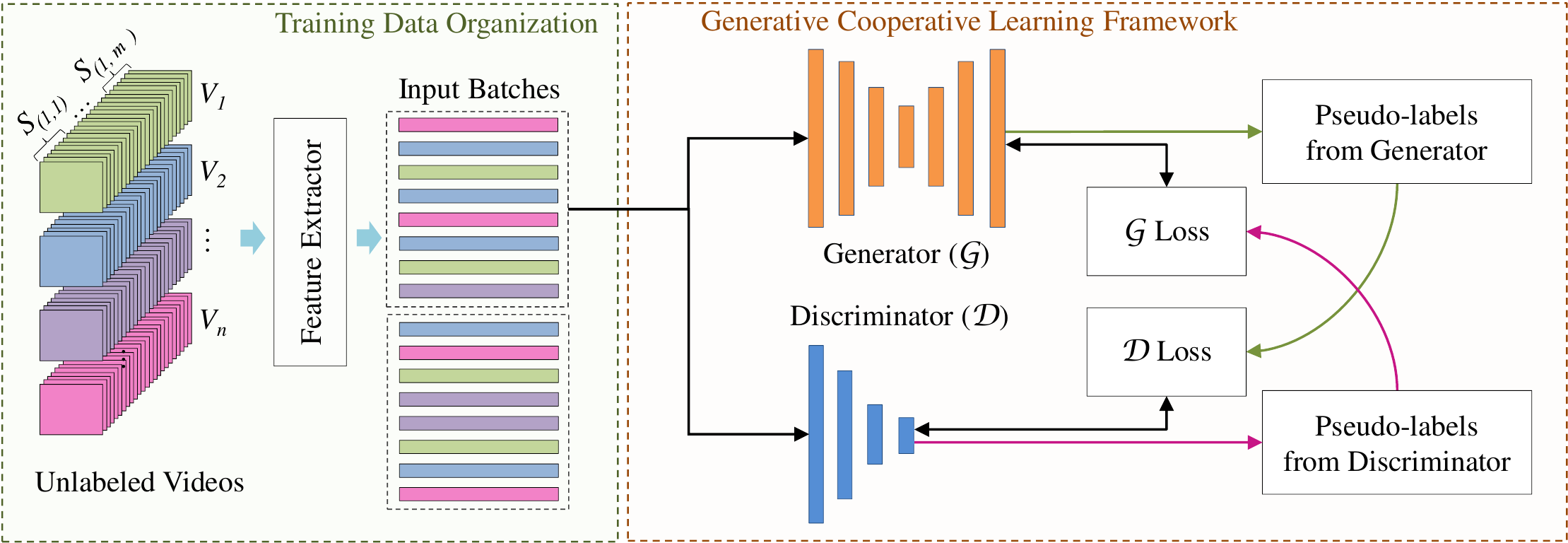}
  \vspace{-1mm}
  \caption{Proposed Generative Cooperative Learning (GCL) algorithm introduces cross-supervision for training a Generator $\mathcal{G}$ and a Discriminator $\mathcal{D}$. The pseudo-labels produced by $\mathcal{G}$ are used to compute the $\mathcal{D}$ loss and likewise, the pseudo-labels produced by the $\mathcal{D}$ are utilized to compute the $\mathcal{G}$ loss. Both $\mathcal{G}$ and $\mathcal{D}$ are trained iteratively from unlabelled training data for anomalous events detection.}
  \label{fig:architecture}
\end{figure*}

\section{Related Work}
Anomaly detection is a widely studied problem both in the image~\cite{chalapathy2019deep,heer2021ood,postels2020hidden} and video domain~\cite{suarez2020survey,zaheer2020old,sultani2018real,zaheer2020claws,zaheer2020self}. We here introduce different supervision modes for video anomaly detection and traditional mutual learning strategies

\noindent \textbf{Anomaly Detection as One-Class Classification (OCC).}
OCC based approaches have found their way in a wide range of anomaly detection problems including, medical diagnosis \cite{wei2018anomaly}, cyber security \cite{gong2019memorizing}, surveillance security systems \cite{zaheer2020old,mohammadi2016angry,kamijo2000traffic,shanghaiTech2017}, and industrial inspection \cite{bergmann2019mvtec}. Some of these approaches use  hand-crafted features \cite{medioni2001event_twostream34,basharat2008learning_realworld7,wang2014learning_realworld38,zhang2009learning_twostream53,piciarelli2008trajectory_twostream36}, while others use deep features extracted using pre-trained models \cite{smeureanu2017deep,ravanbakhsh2017abnormal}. With the advent of generative models, many approaches proposed variants of such networks to learn representations corresponding to normal data \cite{zaheer2020old,Gong_2019_ICCV,ren2015unsupervised,xu2015learning_denoise,Nguyen_2019_ICCV,nguyen2019hybrid,xu2017detecting_denoise,sabokrou2017deep_novelty,sabokrou2020deep}. 
OCC based approaches find it challenging to avoid well-reconstruction of anomalous test inputs. This problem is attributed to the fact that since OCC approaches only use normal class data while training, an ineffective classifier boundary may be achieved which is limited in enclosing normal data while excluding anomalies \cite{zaheer2020old}.
In an attempt to address this limitation, some researchers recently proposed pseudo-supervised methods in which pseudo-anomaly instances are generated using normal training data \cite{zaheer2020old,astrid2021learning}. 

\noindent \textbf{Weakly Supervised (WS) Anomaly Detection.}
Video-level binary annotations are used to train WS classifiers capable of predicting frame-level anomaly scores \cite{sultani2018real,zaheer2020claws,zaheer2020self,zaheer2020cleaning,zhong2019graph,tian2021featuremagnitude,Purwanto_2021_ICCV}. Video-level labels are provided in such a way that a normal labeled video is completely normal whereas an anomalous labeled video contains both normal and anomalous contents without any information about the temporal whereabouts (Fig. \ref{fig:training_modes}). 

\noindent \textbf{Unsupervised Anomaly Detection.}
Anomaly detection methods using unlabelled training data are quite sparse in literature. According to the nomenclature shown in Fig. \ref{fig:training_modes}, most unsupervised methods in the literature actually fall in the category of OCC. For instance, MVTecAD \cite{bergmann2019mvtec} benchmark ensures the training data to be only normal, therefore its evaluation protocol is OCC and the methods inheriting this assumption are also essentially one-class classifiers \cite{Bergmann_2020_CVPR,Gong_2019_ICCV}.
In contrast to these algorithms, our proposed GCL approach is capable of learning from unlabelled training data without assuming any normalcy. The training data in the form of videos conform to several important attributes regarding  anomaly detection, such as, anomalies are less frequent than normal events and events are often temporally consistent.
We derive our motivation from these clues to carry out the training in a completely unsupervised fashion.

\noindent \textbf{Teacher Student Networks.}
Our proposed GCL shares some similarities with the {Teacher Student} (TS) frameworks  for knowledge distillation \cite{hinton2015distilling}. 
GCL is different from TS framework mainly because its aim is not knowledge distillation. Also our generator 
generates noisy labels while our discriminator, being relatively robust to noise, cleans these labels which is not the case in TS framework. 

\noindent \textbf{Mutual Learning (ML).} The GCL framework also shares similarities with the ML algorithms \cite{zhang2018deep}. 
 However, the two components of the GCL learn different types of information and are trained with cross-supervision in contrast to the supervised learning used by the ML algorithms. Further in GCL, the output of each network is passed through a threshold process to produce pseudo-labels. In ML frameworks, the cohort learns to match the distributions of each member while in GCL each member tries to learn from the pseudo-labels generated by the other. A mutual learning of a cohort in unsupervised mode using unlabelled training data is unexplored yet.

\noindent \textbf{Dual Learning.}  It is also a related method in which two language translation models interactively teach each other \cite{he2016dual}. However, the external supervision is provided using pre-trained unconditional language expert models which check the quality of translations.
This way, different models have different learning tasks whereas in our proposed GCL approach the learning tasks are identical.

Another variant of \textbf{Cooperative Learning} \cite{batra2017cooperative} has been previously proposed to learn multiple models jointly for the same task across different domains. 

For instance, object recognition is formulated by training a model on RGB images and another model on depth images which then communicate the domain invariant object attributes. Whereas, in our GCL approach both models address the same task in the same domain.

\section{Method}

Our proposed Generative Cooperative Learning approach for Anomaly Detection (GCL) comprises a feature extractor, a generator network, a discriminator network, and two pseudo-label generators. Fig. \ref{fig:architecture} shows the overall architecture. Each of the components are discussed next.

\subsection{Training Data Organization}
To minimize the computational complexity and to reduce the training time of GCL, similar to the existing SOTA
\cite{zaheer2020claws,zaheer2020cleaning,zaheer2020self,sultani2018real,zhong2019graph,tian2021featuremagnitude}, we also utilize a deep feature extractor to convert video data into compact features. All input videos are arranged as segments, features of which are then extracted. 
Furthermore, these features are randomly arranged as batches.
In each iteration a randomly selected batch is used to train the GCL model (Fig.  \ref{fig:architecture}). Formally, given a training dataset of $n$ videos, every video  is partitioned into non-overlapping segments $S_{(i,j)}$ of $p$ frames each, where $i \in [1, n]$ is the video index 
and $j \in [1, m_i]$ is the segment index. The segment size $p$ is kept the same across all training and test videos of a dataset. 

For each $S_{(i,j)}$, a feature vector $\mathbf{f}_{(i,j)} \in \mathbb{R}^d$ is computed as $\mathbf{f}_{(i,j)}$=$\mathcal{E}(S_{(i, j)})$ using the feature extractor $\mathcal{E}(\cdot)$. 

In the existing weakly supervised anomaly detection approaches, each training iteration is carried out on one or more complete videos \cite{sultani2018real,zhong2019graph}.
Recently, CLAWS Net \cite{zaheer2020claws} proposed to extract several batches of temporally consistent features, each of which was then randomly input to the network. Such configuration serves the purpose of minimizing correlation between consecutive batches. In these existing approaches, it is important to maintain temporal order at batch or video level. However, in the proposed GCL appraoch  we randomize the order of input features which removes both the intra-batch and inter-batch correlations. 

\subsection{Generative Cooperative Learning}
Our proposed Generative Cooperative Learning (GCL) approach for anomaly detection consists of a generator $\mathcal{G}$ which is an autoencoder (AE) and a discriminator $\mathcal{D}$ which is a fully connected (FC) classifier. Both these models are trained in a cooperative fashion without using any data annotations. More specifically, we neither use the normal class annotations as in one class classification (OCC) approaches \cite{Gong_2019_ICCV,wang2019gods,park2020learning}, nor binary annotations used by the weakly supervised anomaly detection systems \cite{zaheer2020claws, zaheer2020self,sultani2018real,zhong2019graph}. As discussed in Section \ref{sec:intro}, the intuition behind using an AE is that such models can somewhat  capture the overall dominant data trends \cite{Gong_2019_ICCV}. On the other hand, the FC  classification network used as a discriminator is known to be efficient when provided with supervised, albeit noisy, training \cite{zaheer2020claws}. In order to carry out the training, first pseudo annotations created using $\mathcal{G}$ are used to train $\mathcal{D}$. In the next step, pseudo annotations created by using $\mathcal{D}$ are used to improve $\mathcal{G}$. Thus, each of the two models are trained by using the annotations created by the other model, in an alternate training fashion. The training configuration aims that the pseudo-labeling is improved over training iterations which consequently results in an improved overall anomaly detection performance. Particular architecture details and several design choices are discussed next.

\subsubsection{Generator Network}

$\mathcal{G}$ takes features as input and produces reconstructions of those features as output.
Typically, $\mathcal{G}$ is trained by minimizing the reconstruction loss $\mathcal{L}_r$ as:
\begin{equation}
\label{eq:recon}
\mathcal{L}_r=\frac{1}{b}\sum_{q=1}^b \mathcal{L}_{G}^q,\text{~}\mathcal{L}_{G}^q=||f^q_{i,j}-\widehat{f}^q_{i,j}||_2,
\end{equation}
where  $f^q_{i,j}$ is a feature vector that is input to $\mathcal{G}$ and $\widehat{f}^q_{i,j}$ is the corresponding reconstructed vector, $b$ is the batch size. 

\subsubsection{Pseudo Labels from Generator}
\label{sec:psuedo_labels_from_generator}
In our proposed collaborative learning, pseudo labels from $\mathcal{G}$ are created to train $\mathcal{D}$. The labels are created by keeping in view the distribution of the reconstruction loss $\mathcal{L}_{G}^q$ of each instance $q$ over a batch. The main idea is to consider feature vectors resulting in higher loss values as anomalous and those generating smaller loss values as normal. In order to implement this intuition, one may consider using a threshold $\mathcal{L}_{G}^{th}$ as:
\begin{equation}
    l^q_G = 
    \begin{cases}
    1, \text{\quad if } \mathcal{L}_{G}^q \ge \mathcal{L}_{G}^{th}\\
    0, \text{\quad otherwise }.
    \end{cases}
\label{eq:g_pseudo}
\end{equation}
We have followed a simple approach for the  $\mathcal{L}_{G}^{th}$ selection by considering a fixed percentage of the samples having maximum reconstruction error as anomalous. In the $\mathcal{L}_{G}^q$ histograms we empirically observed a bigger peak  towards minimum error and  a smaller peak towards maximum error. Due to the fact that the class boundaries often fall in low density regions, error histograms are also an effective tool for the selection of appropriate  $\mathcal{L}_{G}^{th}$. Analysis of different alternates for $\mathcal{L}_{G}^{th}$ selection is given in the Supplementary. 

\subsubsection{Discriminator Network}
The binary classification network used as the discriminator $\mathcal{D}$ is trained using the pseudo annotations from $\mathcal{G}$ by minimizing the binary cross entropy loss over a batch $b$ as: 
\begin{equation}
    \mathcal{L}_D=\frac{-1}{b}\sum_{q=1}^b l^q_G \ln{\widehat{l}^q_{i,j}}+(1-l^q_G) \ln{(1-\widehat{l}^q_{i,j})},   
\end{equation}
where $l^q_G \in \{0, 1\}$ is the pseudo label generated by $\mathcal{G}$ and $\widehat{l}^q_{i,j}$ is the output of $\mathcal{D}$ when a feature vector $f_{i,j}^q$ is input.

\subsubsection{Pseudo Labels from Discriminator}
Pseudo labels from $\mathcal{D}$ are used to improve the reconstruction discrimination capability of $\mathcal{G}$. The output $\widehat{p}_{i,j}^q$ of $\mathcal{D}$ is the probability of a feature vector $f^q_{i,j}$ to be anomalous. Therefore, the features obtaining higher probability are considered as anomalous by using a threshold mechanism on the output $\widehat{p}_{i,j}^q$ of $\mathcal{D}$ . The annotations generated by $\mathcal{D}$ are then used to fine tune $\mathcal{G}$ in the next iteration. 
\begin{equation}
    l^q_D = 
    \begin{cases}
    1, \text{\quad if } \widehat{p}_{i,j}^q \ge \mathcal{L}_{D}^{th}\\
    0, \text{\quad otherwise },
    \end{cases}
\label{eq:d_pseudo}
\end{equation}
where the threshold $\mathcal{L}_{D}^{th}$ is computed the same way as the threshold $\mathcal{L}_{G}^{th}$ is computed.

\begin{figure}[t]
\begin{center}
   \includegraphics[width=.9\linewidth]{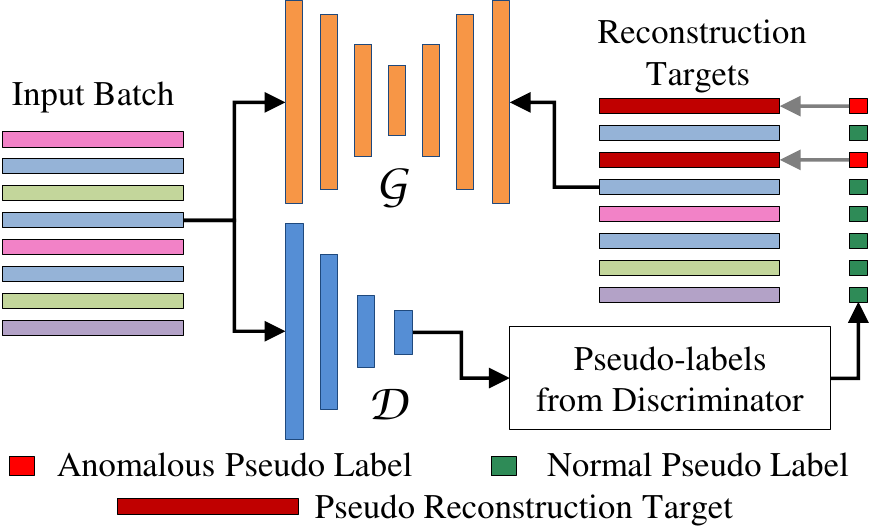}
\end{center} \vspace{-1.5em}
   \caption{Negative learning in GCL: $\mathcal{G}$ is constrained to not learn the reconstruction of anomalies using Pseudo Reconstruction Targets (PRT). Based on the pseudo-labels produced by $\mathcal{D}$, PRT are generated for the anomalous inputs while normal targets are used for the normal inputs to guide the training of $\mathcal{G}$.}
\label{fig:negative_learning_architecture}
\vspace{-1.2em}
\end{figure}

\subsubsection{Negative Learning of Generator Network}
\label{sec:nl_methods}
Training of $\mathcal{G}$ is carried out by using pseudo labels from $\mathcal{D}$ by employing negative learning (NL). In order to increase the discrimination among reconstructions of normal and anomalous inputs, $\mathcal{G}$ is encouraged to poorly reconstruct the samples which have  anomalous pseudo labels whereas, the samples having normal pseudo labels are aimed to be reconstructed as usual with minimum error.

Some variants of NL have already been explored in the literature. For instance, Munawar \etal \cite{munawar2017limiting} and Astrid \etal \cite{astrid2021learning} make the loss negative for a full batch of known anomalous inputs. However, this configuration requires a prior knowledge of the whole dataset and its labels. In the proposed GCL approach, pseudo labels are generated iteratively as the training proceeds, therefore it  may encounter both normal and anomalous samples in the same batch. In addition, instead of making the loss negative, we enforce the abnormal samples to be poorly reconstructed by using a pseudo reconstruction target. Therefore, as illustrated in Fig. \ref{fig:negative_learning_architecture}, for each feature vector which is pseudo-labeled as anomalous by $\mathcal{D}$, its reconstruction target is replaced by a different feature vector. In order to extensively explore this concept, we propose the following different types of pseudo targets:
\textbf{1) All Ones Target}: The original reconstruction target is replaced by a similar dimensional vector of all 1's.
\textbf{2) Random Normal Target}: The original reconstruction target is replaced by a normal labeled feature vector selected arbitrarily. \textbf{3) Random Gaussian Noise Target}: The original reconstruction target is perturbed by adding Gaussian noise. \textbf{4) No Negative Learning}: No negative learning is applied to $\mathcal{G}$. Instead only feature vectors pseudo-labeled as normal are used for the training of $\mathcal{G}$.
Extensive analysis of different pseudo targets is shown in Fig. \ref{fig:nl_techniques_comparison}. We empirically observe that `ones' as pseudo target yields more discriminative  reconstruction  capability, thus better differentiating  normal and anomalous inputs. The loss function given by Eq. \eqref{eq:recon}  is modified to include  negative learning:
\begin{equation}
\mathcal{L}_G=\frac{1}{b}\sum_{q=1}^b||t^q_{i,j}-\widehat{f}^q_{i,j}||_2,    
\end{equation}
where the pseudo target $t_q$ is defined as:
\begin{equation}
    t_{i,j}^q = 
    \begin{cases}
    f_{i,j}^q, \text{\quad if } {l}_D^q  =  {0}\\
    \mathbf{1} \in \mathbb{R}^d, \text{\quad if } {l}_D^q  =  {1},
    \end{cases}
\label{eq:negativelearning}
\end{equation}
\subsection{Self-Supervised Pre-training}
The proposed GCL approach is trained using unlabelled videos by utilizing the cooperation of $\mathcal{G}$ and $\mathcal{D}$. Since anomaly detection is an ill-defined problem,  the lack of constraints may adversely affect the convergence and the system may get stuck in local minima. 
In order to improve the convergence, we explore to \textit{jump-start} the training process by pre-training both $\mathcal{G}$ and $\mathcal{D}$. We empirically observe that using a pre-trained $\mathcal{G}$ (based on Eq. \eqref{eq:recon}) is beneficial for the overall stability of the learning system and it also improves the convergence as well as the performance of the system (see Section \ref{sec:experiments}).

Autoencoders are known to capture dominant representations of the training data \cite{zaheer2020old,Gong_2019_ICCV}. Despite the fact that anomalies are sparse and normal features are abundant in the training data, we experimentally observe that simply utilizing all training data to pre-train $\mathcal{G}$ may not provide an effective jump-start. 
Using the fact that events in videos happen in temporal sequence and that anomalous frames are usually more eventful than the normal ones, we utilize temporal difference between the consecutive feature vectors as an estimator to initially clean the training dataset for the pre-training of $\mathcal{G}$. That is, a feature vector $f_{i,j}^{t+1}$ will only be used for pre-training if $||f_{i,j}^{t+1}-f_{i,j}^t||_2 \le D_{th}$, where the superscripts $t$ and $t+1$ show the temporal order of features in a video and $D_{th}$ is the threshold. This approach does not guarantee  complete removal of anomalous events however, it  cleans  the data for an effective initialization of the $\mathcal{G}$ to give a jump-start to the training. Once $\mathcal{G}$ is pre-trained, it is used to generate psuedo labels which are then used to pre-train the discriminator. In this step, the role of $\mathcal{G}$ is similar to  a lousy teacher because the generated pseudo-labels are quite noisy and the role of $\mathcal{D}$ is like an efficient student because it learns to discriminate normal and anomalous features better even with noisy labels. In the following steps, both pre-trained $\mathcal{G}$ and $\mathcal{D}$ are plugged into our collaborative learning loop.

\subsection{Anomaly Scoring}

In order to compute final anomaly score at test time, several configurations are possible, i.e., using reconstruction error of $\mathcal{G}$ or prediction scores of $\mathcal{D}$. We experimentally observed that $\mathcal{G}$ remains relatively lousy while $\mathcal{D}$ remains  efficient across consecutive training iterations. 
Therefore for simplicity, unless stated otherwise, all results reported in this work are computed using the predictions of $\mathcal{D}$.

\section{Experiments}
\label{sec:experiments}
\noindent In this section, we first provide important experimental details, then draw comparisons with the existing state-of-the-art methods, and finally study different components of our GCL framework.

\noindent \textbf{Datasets.}
UCF-Crime (UCFC) dataset contains 13 different categories of real-world anomalous events which were captured by CCTV surveillance cameras spanning 128 hours \cite{sultani2018real}. This dataset is complex because of the unconstrained backgrounds. The training split contains 810 abnormal and 800 normal videos, while the testing split has 140 anomalous and 150 normal videos. In training split, video-level labels are provided while in test split frame-level binary labels are provided.
In our unsupervised setting, we discard the training-split labels and train the proposed GCL using unlabelled training videos. 

ShanghaiTech contains staged anomalous events captured in a university campus at 13 different locations spanning 437 videos. This dataset was originally proposed for OCC with only normal videos provided for training. Later, Zhong \etal \cite{zhong2019graph} reorganized this dataset to facilitate training of weakly-supervised algorithms. Normal and anomalous videos were mixed in both the training and the testing splits. The new training split contains 63 anomalous and 175 normal videos whereas, the new testing split contains 44 anomalous and 155 normal videos. In order to train our proposed GCL, we follow the latter split both for training and testing, without using training split video labels.

\noindent\textbf{Evaluation Measures.}
Following the existing methods \cite{lu2013abnormal,zhong2019graph,sultani2018real,hasan2016anomaly}, we use area under ROC curve (AUC) for evaluation and comparisons. AUC is computed based on frame-level annotations of the test videos in both datasets.

\begin{figure}[t]
\begin{center}
     
        \includegraphics[width=1\linewidth]{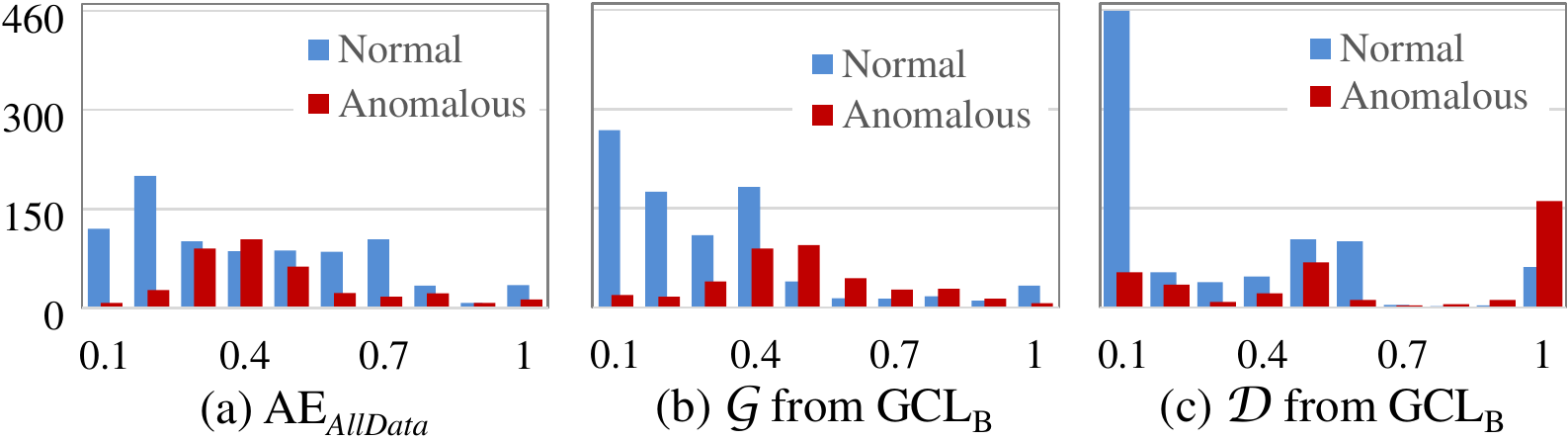}
\end{center}
\vspace{-1.5em}
     \caption{Distribution of scores predicted on the test split of UCF-crime dataset by (a) $\mathcal{AE}$ trained on all training data, (b) $\mathcal{G}$ trained in GCL$_B$, and (c) $\mathcal{D}$ trained in GCL$_B$. Although $\mathcal{G}$ and $\mathcal{D}$ are trained cooperatively, $\mathcal{D}$ being more robust to noise, demonstrates superior discrimination between normal and anomalous examples.}
     \label{fig:histogram analysis}
   \end{figure}

\noindent\textbf{Implementation Details.}
To demonstrate the concept of cooperative learning in its true essence, we select fairly simple architectures, without any bells and whistles, as our $\mathcal{G}$ and $\mathcal{D}$ networks. Architectures of $\mathcal{G}$ and $\mathcal{D}$ are set as FC[2048, 1024, 512, 256, 512, 1024, 2048] and FC[2048, 512, 32, 1]. We train both networks using RMSprop optimizer with a learning rate of 0.00002, momentum 0.60, for 15 epochs on training data with batch size $8192$. Thresholds for pseudo-label generation are data driven. For $\mathcal{G}$ pseudo-labels $\mathcal{L}_{\mathcal{G}}^{th}=\mu_R+\sigma_R$ where $\mu_R$ and $\sigma_R$ are the mean and the standard deviation of reconstruction error as given by Eq. \eqref{eq:recon} for each batch. For $\mathcal{D}$, $\mathcal{L}_{D}^{th}=\mu_P+0.1 \sigma_P$, where $\mu_P$ and $\sigma_P$ are the mean and standard deviations of the probabilities $\widehat{p}_{i,j}^q$ generated by $D$ for each batch. The value of $D_{th}$=0.70 is used in unsupervised pre-training. As feature extractor, we use a popular framework ResNext3d proposed by Hara \etal \cite{hara3dcnns} in default mode. Segment size $p$ for feature extraction is set to 16 non-overlapping frames. All experiments are performed on NVIDIA RTX 2070 with Intel Core i7, 8th gen and 16GB RAM. Code will be released upon acceptance.

\begin{table}[]
\begin{center}
\caption{Performance comparison with existing state-of-the-art methods on UCF-Crime (UCFC) and ShanghaiTech (STech) datasets.  We divide the methods into three categories based on the supervision used in training. Best results are in bold.}
\label{tab:UCF_crime_AUC}
\resizebox{\columnwidth}{!}{
\begin{tabular}{c|c|c|c|c}
{\makecell{\textbf{Supervision}\\\textbf{Type}}}                                     & \textbf{Method}     &\textbf{\revised{Features}}                     & {\makecell{\textbf{UCFC}\\\textbf{AUC\%}}}  & {\makecell{\textbf{STech}\\\textbf{AUC\%}}}  \\ \hline\hline
\multirow{8}{*} {\makecell{\textbf{One Class}\\\textbf{Classification}}}  & SVM \cite{sultani2018real}   &  -                      &   50.00   & -   \\\cline{2-5} 
                                          & Hasan \etal \cite{hasan2016anomaly}           &  -                        &  50.60   & 60.85   \\\cline{2-5} 
                                          & Sohrab \etal \cite{sohrab2018subspace}        &  -                  &  58.50  & -    \\\cline{2-5} 
                                          & Lu \etal \cite{lu2013abnormal}                &  -               &  65.51  & 68.00    \\\cline{2-5}
                                          & BODS \cite{wang2019gods}                      &  I3D               &  68.26  & -    \\\cline{2-5}
                                          & OGNet**  \cite{zaheer2020old}                &  ResNext               &  69.47 & 69.90      \\\cline{2-5}
                                          & GODS  \cite{wang2019gods}                     &  I3D                &  70.46 & -     \\ \cline{2-5}
                                          & TSC  \cite{luo2017revisit}                    &  -                 &  - & 67.94     \\ \cline{2-5}
                                          & Frame Prediction  \cite{liu2018future}        &  -                             &  - & 73.40     \\ \cline{2-5}
                                          & MemAE  \cite{gong2019memorizing}              &  -                       &  - & 71.20     \\ \cline{2-5}
                                          & MNAD  \cite{park2020learning}                 &  -                    &  - & 70.50     \\ \cline{2-5}
                                          & Cho \etal  \cite{cho2020unsupervised}                            &  -                    &  - & 74.70     \\ \cline{2-5}
                                          & LNTR  \cite{astrid2021learning}               &  -                      &  - & 75.97     \\ \cline{2-5}
                                          & RUVAD  \cite{wang2021robust}                                &  -                    &  - & 76.67     \\ \cline{2-5}
                                          & BMAN  \cite{lee2019bman}                      &  -               &  - & 76.20     \\ \cline{2-5}
                                          & \textit{Proposed}  GCL$_{OCC}$                &  ResNext                     &  \textbf{74.20} & \textbf{79.62$^*$}     \\ \hline\hline
\multirow{11}{*}{\makecell{\textbf{Weak}\\\textbf{Supervision}}}       & Sultani \etal \cite{sultani2018real}     &    C3D   &  75.41   & -   \\\cline{2-5} 
                                          & Zhang \etal \cite{zhang2019temporal}          &  C3D                    &  78.66  & 82.50    \\\cline{2-5} 
                                          & Zhu \etal \cite{zhu2019motion}                &  C3D            &  79.00  & -    \\\cline{2-5} 
                                          & Noise Cleaner \cite{zaheer2020cleaning}       &  C3D                  &  78.27  & 84.16    \\\cline{2-5} 
                                          & SRF \cite{zaheer2020self}                     &  C3D              &  79.54  & 84.16    \\\cline{2-5}
                                          & DUAD*** \cite{li2021deep}                                  &  C3D              &  72.90  & -    \\\cline{2-5}
                                          & GCN \cite{zhong2019graph}                     &  C3D                &  81.08  &  76.44    \\\cline{2-5} 
                                          & GCN \cite{zhong2019graph}                     &  TSN$^{RGB}$               &  82.12  & 84.44    \\\cline{2-5} 
                                          & Wu \etal \cite{wu2020not}                     &  I3D           &  82.44  & -    \\\cline{2-5} 
                                          & DAM \cite{majhidam}                           &  I3D         &  82.67   &  88.22  \\\cline{2-5} 
                                          & CLAWS  \cite{zaheer2020claws}                 &  C3D                 &  83.03   & 89.67   \\\cline{2-5} 
                                          & CLAWS** \cite{zaheer2020claws}                &  \revised{ResNext}                 &  82.61   & -   \\\cline{2-5} 
                                          & Yu \etal  \cite{tian2021featuremagnitude}     &  C3D                 &  83.28 & 91.51     \\\cline{2-5} 
                                          & Yu \etal  \cite{tian2021featuremagnitude}     &  I3D                  &  84.30 & \textbf{97.27}     \\\cline{2-5} 
                                          & Purwantu \etal  \cite{Purwanto_2021_ICCV}     &  TRN                     &  \textbf{85.00}  & 96.85    \\\cline{2-5} 
                                          & \textit{Proposed} GCL$_\text{WS}$             &  ResNext         &  79.84  & 86.21    \\ \hline\hline
\multirow{4}{*}{\textbf{Unsupervised}}    & kim \etal** \cite{kim2021semi}                &  ResNext         &  52.00  & 56.47     \\\cline{2-5} 

                                          & AE$_\text{AllData}$                           &  ResNext     &  56.32 & 62.73     \\\cline{2-5} 
                                          & \textit{Proposed} GCL$_\text{B}$              &  ResNext    &  68.17  & 72.41      \\\cline{2-5} 
                                          & \textit{Proposed} GCL$_\text{PT}$             &  \revised{C3D}    &  70.74  & -      \\\cline{2-5} 
                                          & \textit{Proposed}  GCL$_\text{PT}$            &  ResNext    &  \textbf{71.04}  & \textbf{78.93}    \\\hline 
\end{tabular}

}
$^*$ We follow the evaluation protocol of Zhong \etal \cite{zhong2019graph}. \revised{$^{**}$We implemented the models and computed these scores. ***\cite{li2021deep} computes scores by taking average over videos.}
\vspace{-2.5em}
\end{center}
\end{table}

\subsection{Comparisons with State-Of-The-Art (SOTA)}
The proposed GCL approch is trained in an unsupervised fashion without using any video-level or frame-level annotations. GCL  with no pre-training, GCL$_{B}$, is considered as the baseline. In addition, GCL with pre-training, GCL$_{PT}$, GCL combined with OCC based pre-trained autoencoder, GCL$_{OCC}$, and GCL weakly-supervised, GCL$_{WS}$ are also trained and evaluated on UCFC and ShanghaiTech datasets.  

As seen in Table \ref{tab:UCF_crime_AUC}, on \textbf{UCFC dataset}, the proposed GCL$_{B}$ obtained an overall AUC of 68.17 \% which is 11.85 \% higher than the Autoecnoder (AE$_{AllData}$) trained on complete training data including both normal and anomalous training samples in an unsupervised fashion. Histogram plotted over reconstructions in Fig. \ref{fig:histogram analysis}(a) also provides insights that AE$_{AllData}$ is not able to learn discriminative reconstruction. Also in the GCL, the discrimination ability of $\mathcal{D}$ (Fig. \ref{fig:histogram analysis}(c)) is much enhanced than $\mathcal{G}$ (Fig. \ref{fig:histogram analysis}(b)). Experiments on kim \etal \cite{kim2021semi} are conducted on a re-implementation of the method for unlabelled training data.

GCL$_{PT}$ is the version of proposed GCL with an autoencoder pre-trained in an unsupervised fashion. In this experiment, an AUC performance of  71.04\% is obtained which is 2.87\% better than the baseline GCL$_B$. \revised{ The two methods are also compared in Fig. \ref{fig:convergence} using multiple random seed initialization and GCL$_{PT}$ demonstrates consistent performance gains.}
Table \ref{tab:UCF_crime_AUC} also shows that the proposed GCL$_{PT}$ outperforms all existing one-class classification based anomaly detection methods. It is despite the fact that while training  GCL$_{PT}$, no labeled supervision is used. In contrast, OCC methods use clean normal class for training which provides extra information compared to our unsupervised training based GCL.

In another experiment, the autoencoder is pre-trained on only the normal class of the training data, which makes the setting comparable with the one-class classifiers. This scheme of extra information provided in the form of normal class labels, referred as GCL$_{OCC}$ in Table \ref{tab:UCF_crime_AUC}, obtains an improved performance of 74.20\% on UCFC which is significantly better than all existing state-of-the-art OCC methods. 
It is interesting to note that GCL$_{OCC}$ yields comparable performance to the approach proposed by Sultani \etal \cite{sultani2018real} which utilizes video-level labels for training. 

Although GCL aims at unsupervised cooperative learning, we also extended it to incorporate weak-supervision. The results for this version are reported as GCL$_{WS}$ in Table \ref{tab:UCF_crime_AUC}. Despite using fairly simple networks of $\mathcal{G}$ and $\mathcal{D}$ without any bells and whistles, GCL$_{WS}$ obtains comparable results to several existing weakly-supervised learning methods.

\begin{figure}[t]
\begin{center}
    \subfloat[Training of $\mathcal{G}$ in GCL$_{B}$]
    {
       \includegraphics[width=.48\linewidth]{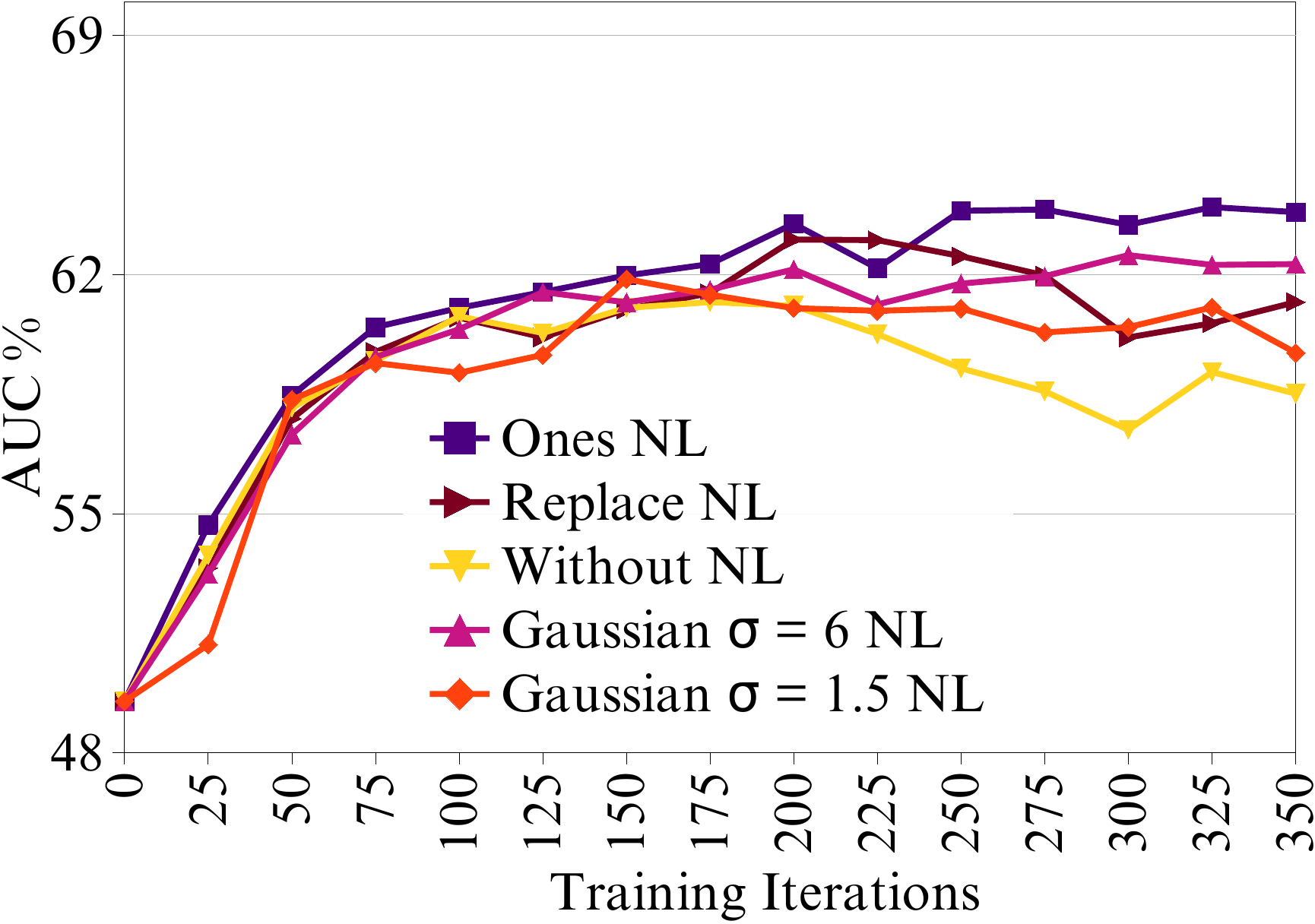}
       
    }
     \subfloat[Training of $\mathcal{D}$ in GCL$_{B}$]
     {
        \includegraphics[width=.48\linewidth]{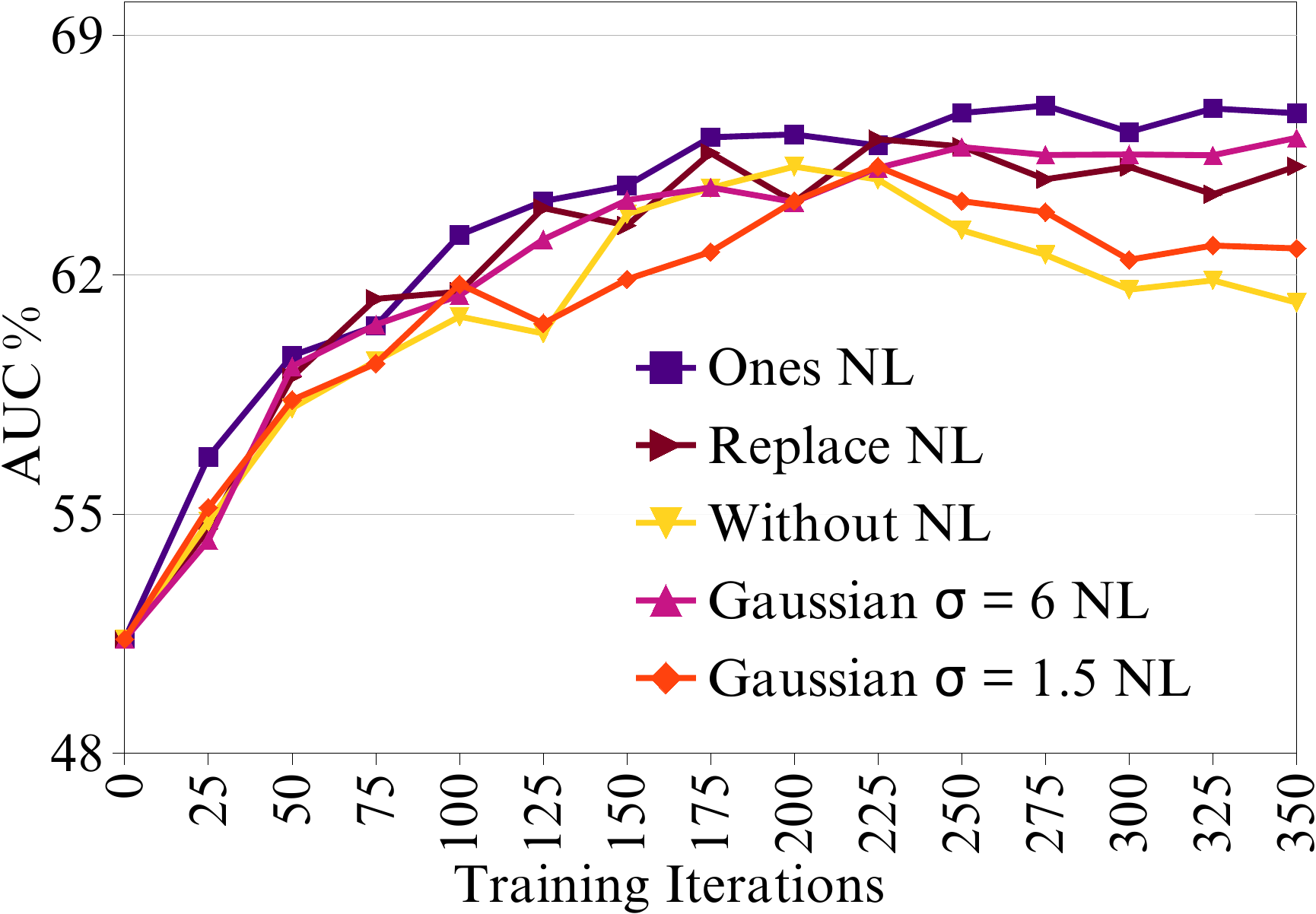}
       
     }
     \vspace{-1.5em}
\end{center}
     \caption{Convergence of $\mathcal{G}$ and $\mathcal{D}$ in GCL with/without Negative Learning (NL). We test different pseudo reconstruction targets in NL. Best performance is observed for `ones' NL target. 
     }
     \label{fig:nl_techniques_comparison}
   \end{figure}

We also evaluated our approach on \textbf{ShanghaiTech dataset} \cite{shanghaiTech2017} and the results are compared with the existing SOTA methods in Table \ref{tab:UCF_crime_AUC}. On this dataset, our proposed GCL$_B$ obtained 72.41\% AUC which is more than 10\% better than AE$_{AllData}$ showing the effectiveness of the baseline approach.  GCL$_{PT}$ obtained 78.93\% AUC which is 6.5\% better than GCL$_{B}$ demonstrating the importance of unsupervised pre-training for jump-start. Also, although unsupervised, GCL$_{PT}$ outperformed all existing OCC methods. The experiments on ShanghaiTech dataset also demonstrate the effectiveness of the proposed GCL$_B$ and GCL$_{PT}$ algorithms for anomalous events detection using unlabelled training data.

\begin{figure}[b]
\begin{center}
    \subfloat[AE$_{AllData}$]
    {
       \includegraphics[width=.30\linewidth]{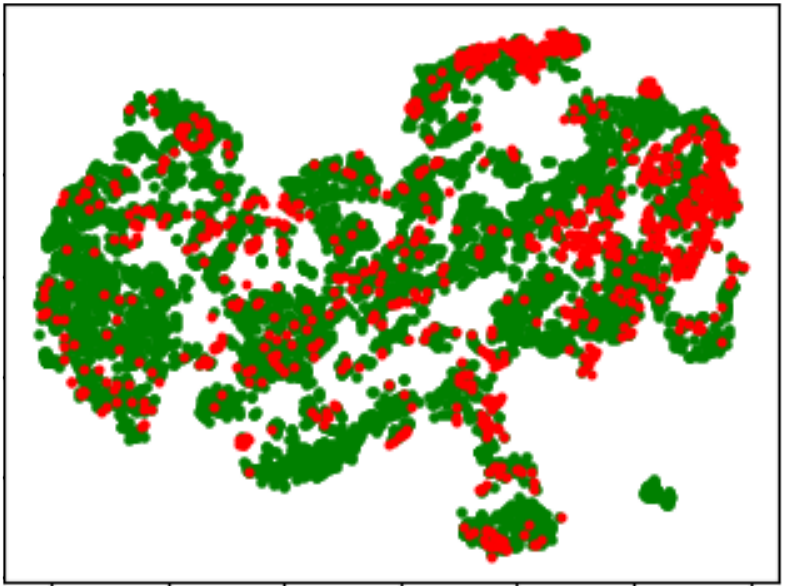}
       
    }
     \subfloat[AE in GCL$_{w/oNL}$ ]
     {
        \includegraphics[width=.30\linewidth]{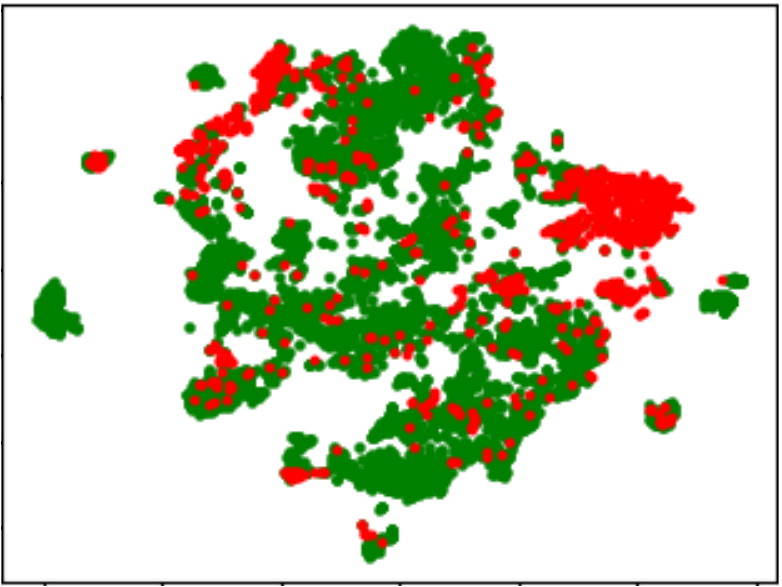}
     }
    \subfloat[AE in GCL$_{B}$ ]
    {
       \includegraphics[width=.30\linewidth]{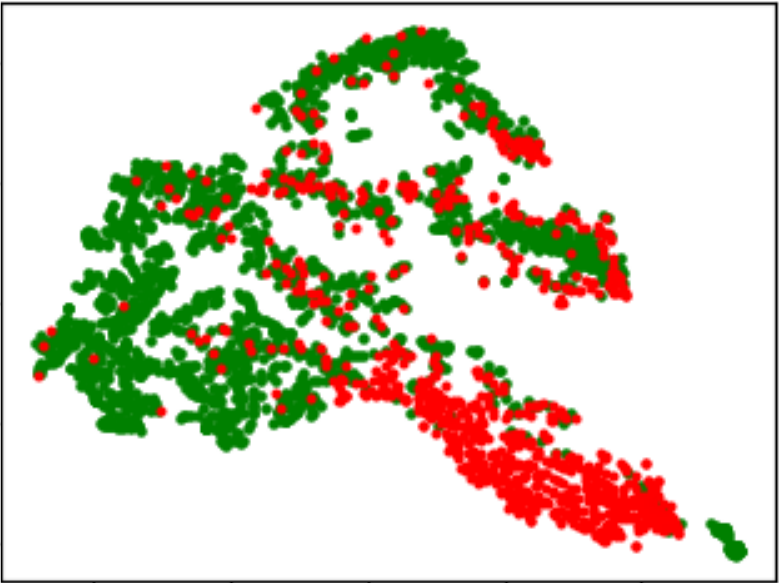}
    }
\end{center} \vspace{-1.5em}
     \caption{Feature reconstructions, using tSNE, with a) AE$_{AllData}$, b) AE in GCL$_{w/oNL}$ without NL, and c) AE in GCL$_B$ with NL using `ones' pseudo targets. Red and green points represent ground truth anomalous and normal samples, respectively. Using negative learning (NL), most of the anomalous samples get clustered separately from the normal samples, which is the underlying desideratum of providing pseudo reconstruction targets.}
     \label{fig:tsne_nl}
   \end{figure}

\begin{figure*}[t]
\begin{center}
   \includegraphics[width=1\linewidth]{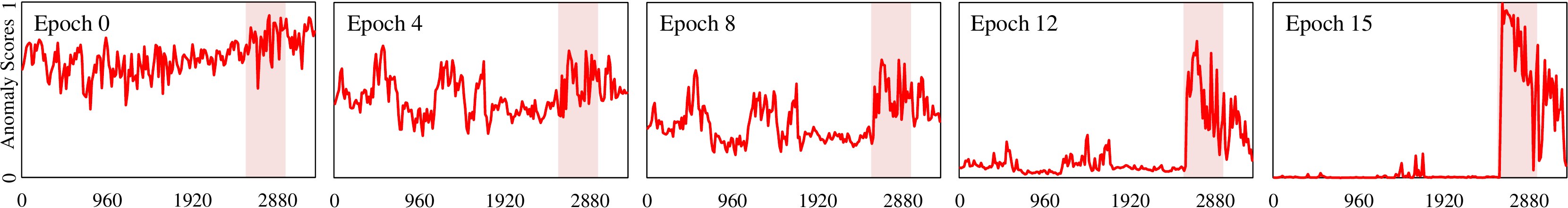}
\end{center}
     \vspace{-4mm}
   \caption{Evolution of the frame-level anomaly scores in GCL$_B$ framework during training. Note that our unsupervised approach successfully produces significantly higher scores in the anomalous portions whereas lower scores in the normal portions. Anomaly ground truth is shown as red boxes, and the video is \textit{Explosion013} from UCF-Crime. Interestingly, the anomaly score stays higher after the anomalous ground truth is over which is essentially due to aftermath of the explosion that network figures to be anomalous.}
\label{fig:scores_evolution}

\vspace{-3mm}
\end{figure*}

\begin{figure*}[t]
\begin{center}
   \includegraphics[width=1\linewidth]{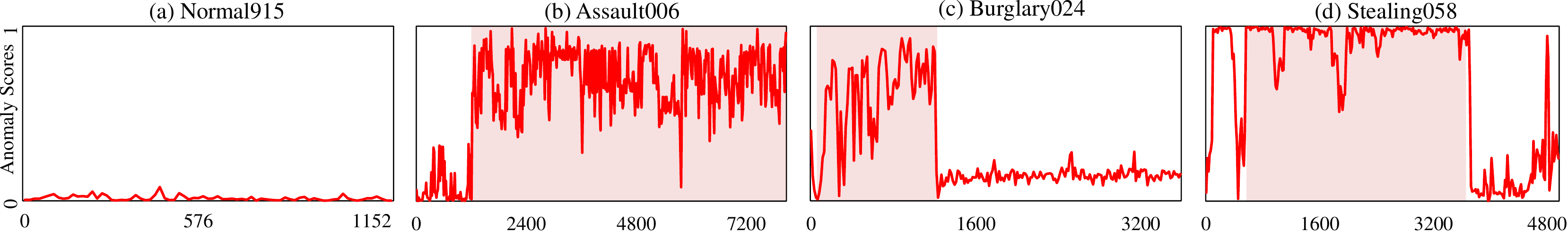}
\end{center}
\vspace{-1.5em}
   \caption{Anomaly scores by GCL$_{PT}$ are low in normal regions and high in abnormal regions on four different UCFC videos.} 
\label{fig:qualitative_results}
\vspace{-1.5em}
\end{figure*}

\subsection{Ablation Study and Analysis} 
\noindent Analysis of different components, design choices, qualitative results and inclusion of supervision are discussed next.

\noindent\textbf{Component-wise ablation study.} A detailed ablation analysis of GCL framework  with various design choices is reported in Table \ref{tab:ablation} on the UCFC dataset.
It can be seen that an autoencoder trained using all training dataset without any supervision AE$_{AllData}$ yields a significantly low performance of 56.32\% compared to the one trained on clean normal data in OCC setting AE$_{OCC}$ yielding AUC of 65.76\%. Training an autoencoder AE$_{TD}$ with our proposed frame temporal difference based unsupervised pre-processing brings the performance closer to AE$_{OCC}$, which demonstrates the superiority of our proposed pre-processing approach. Using negative learning enhances the overall performance of GCL$_B$ over the counterpart training without negative learning GCL$_{w/oNL}$ by 3.94\%. Our complete unsupervised system GCL$_{PT}$ which utilizes negative learning and unsupervised pre-training enhances the overall performance to 71.04\%. In addition, in GCL$_{OCC}$ adding one-class supervision improves this performance even further by demonstrating an AUC of 74.20\%. This also re-validates our claim of the overall benefit that OCC may have over a completely unsupervised setting, making them different from the unsupervised approaches.

\begin{table}[t]
\begin{center}
\vspace{-2mm}
\caption{Ablation analysis of GCL Algorithm: performance of different components with varying supervision levels. } \vspace{-1em}
\label{tab:ablation}
\resizebox{\columnwidth}{!}{
\begin{tabular}{c|cc|c|c|c}
            & \multicolumn{2}{c|}{\textbf{Supervision}}        & \multirow{2}{*}{\makecell{\textbf{Negative}\\\textbf{learning}}} & \multirow{2}{*}{\makecell{\textbf{Unsup.}\\\textbf{pre-training}}} & \multirow{2}{*}{\textbf{AUC\%}} \\ \cline{2-3}
            & \multicolumn{1}{c|}{\textbf{OCC}} & \textbf{Unsup.} &                                    &                                            &                         \\ \hline\hline
AE$_{AllData}$ & \multicolumn{1}{c|}{-}   & \cmark            & -                                  & -                                          & 56.32                   \\ \hline
AE$_{OCC}$ & \multicolumn{1}{c|}{\cmark}   & -            & -                                  & -                                          & 65.76                   \\ \hline
AE$_{TD}$ & \multicolumn{1}{c|}{-}   & -            & -                                  & \cmark                                          & 63.84                   \\ \hline
GCL$_{w/oNL}$        & \multicolumn{1}{c|}{-}   & \cmark            & -                                  & -                                          & 64.23                   \\ \hline
GCL$_{B}$         & \multicolumn{1}{c|}{-}   & \cmark            & \cmark                                  & -                                          & 68.17                   \\ \hline
GCL$_{PT}$        & \multicolumn{1}{c|}{-}   & \cmark            & \cmark                                  & \cmark                                          & 71.04                   \\ \hline
GCL$_{OCC}$       & \multicolumn{1}{c|}{\cmark}   & -            & \cmark                                  & \cmark                                          & 74.20                  
\end{tabular}
}
\end{center}
\end{table}

\noindent\textbf{Evaluating negative learning approaches.} 
Experiments are performed with and without Negative Learning (NL) with GCL framework on UCFC dataset. For the case of NL, GCL$_{B}$, the performances of different pseudo targets (discussed in section \ref{sec:nl_methods}) are also compared in Fig.  \ref{fig:nl_techniques_comparison}. Three different types of pseudo targets are compared: `ones' for all ones, `replace' with random normal, and `gaussian' with $\mu=0$ and $\sigma=\{1.5, 6.0\}$.
Fig. \ref{fig:nl_techniques_comparison} shows that the `ones' pseudo target  works better than its counterpart approaches. Gaussian perturbations with $\sigma=1.5$ demonstrate almost identical trend to the model without any negative learning GCL$_{w/oNL}$, and with $\sigma=6$ the performance improves but still lower than the `ones' performance, GCL$_{B}$. This could be that the fixed pseudo-target helps consistent learning of GCL framework resulting in better discrimination.

To further explore the significance of NL, we provide tSNE visualizations of the reconstructions produced by AE$_{AllData}$, GCL$_{w/oNL}$ AE without NL, and GCL$_B$ AE with NL (trained using `Ones' pseudo label) in Fig. \ref{fig:tsne_nl}. AE$_{AllData}$  is trained using all training data without any labels.
Both GCL$_B$ AEs with and without NL demonstrate a superior discrimination over AE$_{AllData}$. Moreover, in AE with NL (Fig. \ref{fig:tsne_nl}(c)), the anomalous features are forming a distinct cluster which shows that the use of NL with pseudo  reconstruction target is effective than using no NL option.

\begin{figure}[b]
\begin{center}
  \includegraphics[width=0.7\linewidth]{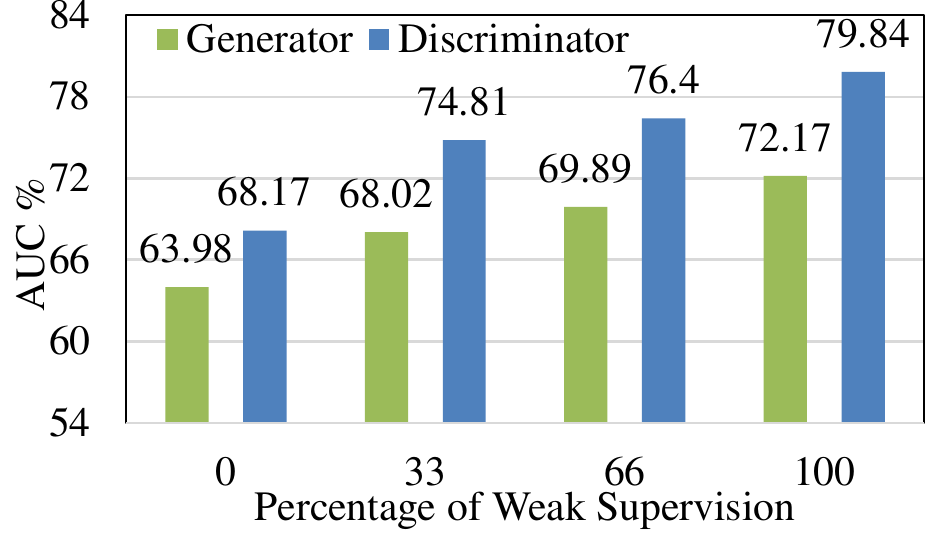}
\end{center}
\vspace{-1.5em}
  \caption{Performance evaluation of $\mathcal{G}$ and $\mathcal{D}$ in weakly supervised GCL$_{WS}$ by increasing supervision level from 0 to 100\%. 
  }
\label{fig:adding_supervision}
\end{figure}

\noindent\textbf{Qualitative analysis.}
A step by step evolution of our GCL approach is visualized in Fig. \ref{fig:scores_evolution}. As the training proceeds, GCL$_B$ learns to predict true anomalous portions within the video in a completely unsupervised fashion. Fig. \ref{fig:qualitative_results} shows final anomaly scores predicted by our GCL$_{PT}$ on four different videos taken from UCFC dataset. 
In Fig. \ref{fig:qualitative_results}(d), some normal portions are also predicted as anomalous. A visual inspection of this video reveals that the beginning and ending frames contain floating text, which is unusual in the training data.

\begin{figure}[b]
\begin{center}
   \includegraphics[width=1\linewidth]{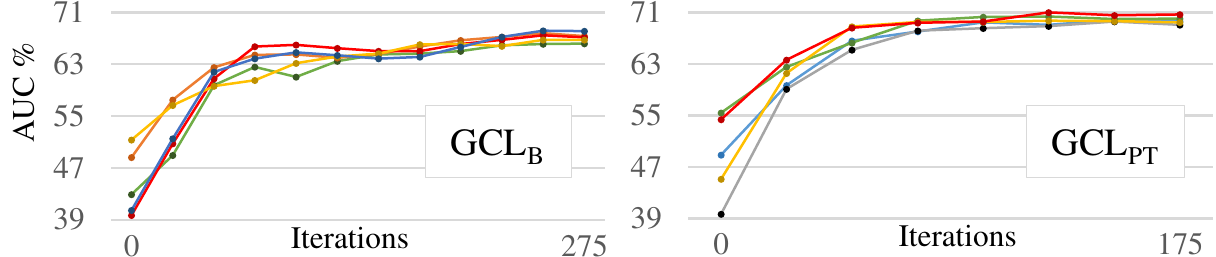}
\end{center}
     \vspace{-4mm}
   \caption{\revised{Convergence of both $GCL_{B}$ and $GCL_{PT}$ by initiating training using several random seeds. $GCL_{B}$ and $GCL_{PT}$ obtain average AUC of $67.09\pm0.65$ and $70.13\pm0.52$ respectively. $GCL_{PT}$ not only improves the overall performance but also reduces the variation over different seeds, thereby demonstrating better convergence.
 }}
\label{fig:convergence}
\end{figure}

\noindent\textbf{On convergence.} \revised{We (empirically) validate the convergence of both GCL$_{B}$ and GCL$_{PT}$ using multiple (10) random seed initialization in Fig.~\ref{fig:convergence}. $GCL_{B}$ and $GCL_{PT}$ obtain an average AUC of $67.09\pm0.65$ and $70.13\pm0.52$, respectively. $GCL_{PT}$ not only improves the overall performance but also reduces the variation over different seeds, thereby demonstrating better convergence.}

\noindent\textbf{On adding weak-supervision.}
In a series of experiments using UCFC, weak video-level labels are infused to the GCL ranging from 33\% to 100\%. Fig.  \ref{fig:adding_supervision} demonstrates that both $\mathcal{G}$ and $\mathcal{D}$ benefit from the added supervision. Noticeably, there is a significant jump in AUC\% upon only providing 33\% videos with weak labels which demonstrates the fact that even minimal supervision is quite beneficial for the proposed GCL. 

\noindent\textbf{\revised{On training $\mathcal{G}$ using its own pseudo-labels.}}
\revised{To further understand proposed collaborative training, we also explore a possibility of training $\mathcal{G}$ using its own pseudo-labels. We employ negative learning to generate labels for training of $\mathcal{G}$ using the reconstruction error of $\mathcal{G}$ itself. Under this configuration, we observed a performance of 62.28\% on UCF crime dataset using ResNext3d features. It is better than 56.32\%, the performance of $AE_{AllData}$ (Table \ref{tab:UCF_crime_AUC}), however noticeably lower than 71.04\%, the performance of our proposed $GCL_{PT}$. This demonstrates that the usage of $\mathcal{D}$ for pseudo-labeling is critical due to its robust learning under
noisy labels \cite{zaheer2020claws,zaheer2020self}. Since $\mathcal{G}$ creates noisy pseudo-labels, $\mathcal{D}$ being robust to noise effectively cleans these labels ensuring the success of the overall collaborative learning.}

\noindent\textbf{\revised{On using soft labels.}}
\revised{In our current configuration, while using pseudo-labels of $\mathcal{G}$ to train $\mathcal{D}$, a threshold is applied to create binary labels from the reconstruction error (eq. \eqref{eq:g_pseudo}). However, it is also possible  that we use soft labels instead of thresholding. Carrying out this experiment on UCF crime dataset using ResNext3d features resulted in a AUC of 63.58\%. Interestingly, the performance is almost identical to that of $AE_{TD}$ in Table \ref{tab:ablation}. Intuitively, it is because without threshold, $\mathcal{D}$ simply starts replicating the output of $\mathcal{G}$, thereby demonstrating identical performance.}

\noindent\textbf{Limitations.}
The proposed unsupervised setting enables an anomaly detection system to start detecting abnormalities just based on the observed data without any human intervention. In case there is no abnormal event so far, the system may consider the rare normal events as abnormal. However, if a system remains operational for a significant time, the probability of having no abnormal event will be rather very small.

\vspace{-2mm}
\section{Conclusion}\vspace{-1mm}
We proposed an unsupervised anomaly detection approach (GCL) using unlabeled training videos, which can be deployed without providing any manual annotations. 
GCL shows excellent performance on two public benchmark datasets with varying supervision levels, including no-supervision, one class and weak-supervision. 
Finally, we discussed the limitations of unsupervised settings, i.e., the assumption of having anomalies in the training dataset. However, this is more realistic than OCC methods as it is natural to have anomalies in the real-world scenarios.

\section{Acknowledgements}
This work was supported by the seed-type challenge research project grant funded by Electronics and Telecommunications Research Institute (ETRI) (No. 21YS2700, Development of learning model and data generation/augmentation techniques for data efficient deep learning, 50\%) and also supported by ETRI with a grant funded by Ulsan Metropolitan City (22AS1600, the development of intelligentization technology for the main industry for manufacturing innovation  and Human-mobile-space autonomous collaboration intelligence technology development in industrial sites, 50\%)

%-------------------------------------------------------------------------

{\small
\bibliographystyle{ieee_fullname}
\bibliography{PaperForReview}
}

\end{document}